\newif\ifarxiv
\arxivtrue    

\ifarxiv
  \documentclass[sigconf]{acmart}
\else
  \documentclass[sigconf,anonymous]{acmart}
\fi

\renewcommand\footnotetextcopyrightpermission[1]{}
\usepackage{tikz}
\usetikzlibrary{positioning, arrows.meta, calc}
\usepackage{float}
\usepackage{stfloats} 
\usepackage{pgfplots}
\usepackage{placeins}
\usepackage{booktabs}
\pgfplotsset{compat=1.18}

\begin{document}
\title{Enforcing Monotonic Progress in Legal Cross-Examination:\\ Preventing Long-Horizon Stagnation in LLM-Based Inquiry}
\subtitle{This paper is under review for ICAIL 2026.}

\ifarxiv
\author{LIAO, Hsien-Jyh}
\affiliation{
  \country{Taiwan, ROC}
}
\email{hjliao12345@gmail.com}
\fi

\begin{abstract}
Large language models (LLMs) exhibit impressive linguistic fluency but struggle to
reliably complete long-horizon tasks under explicit procedural constraints.
In legal cross-examination, purely probabilistic generation often maintains
behavioral coherence while failing to ensure procedural advancement.
We characterize this failure as \emph{procedural stagnation} and propose
\textbf{Soft-FSM}, a neuro-symbolic architecture that enforces monotonic progress
over accumulated Key Information Units (KIUs) via an external deterministic
state controller.
Experiments on three real-world Taiwanese criminal homicide cases show that
baseline methods collapse below 40\% completeness, while Soft-FSM consistently
achieves over 97\% with near-zero redundancy.
These results suggest that, in such domains, reliable task completion cannot be guaranteed by emergent LLM behavior alone, and can be reliably enforced through explicit and verifiable external state control.

\end{abstract}


\maketitle
\thispagestyle{plain}
\pagestyle{plain}
\section{Introduction} 

Automated legal interrogation represents a paradigm shift in legal AI, moving from
passive information retrieval to active, truth-seeking inquiry.
In such systems, an inquirer must conduct multi-turn interactions to incrementally
elicit structured factual elements required to reconstruct complex case narratives.
Unlike open-ended dialogue, legal cross-examination exhibits pronounced role
asymmetry: the inquirer bears global strategic responsibility—including focus
selection, stage transitions, and termination decisions—while the respondent remains
purely reactive.

Recent surveys on computational argumentation and dialogue systems identify
long-horizon logical inconsistency as a persistent open challenge~\cite{castagna2024}.
Although techniques such as equilibria prompting and self-correction can mitigate
surface-level context drift in large language models (LLMs)~\cite{drift2025}, these
approaches remain insufficient in asymmetric, procedurally constrained settings.
Here, even minor per-turn procedural misjudgments can accumulate over long horizons,
leading to structural collapse despite preserved linguistic coherence—a phenomenon
we term the \textbf{Complexity Cliff}.
With per-turn error rate $\epsilon$, the probability of eventual failure exceeds
$1 - (1 - \epsilon)^n$ over $n$ turns, rendering reliable completion increasingly difficult to guarantee as task depth grows.

To address this limitation, we propose \textbf{Soft-FSM}, a neuro-symbolic architecture
that decouples procedural state management from natural language generation.
Soft-FSM operationalizes an external deterministic finite-state controller to enforce
monotonic advancement over accumulated \emph{Key Information Units} (KIUs), permitting
stage transitions only when verifiable information gain is achieved
(i.e., $|K_{t+1}| > |K_t|$).
This design enforces progress-sensitive inquiry without sacrificing the linguistic
flexibility of LLMs for question formulation.

Experiments on three real-world Taiwanese criminal homicide cases—ranging from simple
confession to complex denial involving multiple accomplices—demonstrate the
effectiveness of this approach.
While pure LLM and stage-prompted baselines collapse below 40\% completeness in
challenging settings, Soft-FSM consistently achieves over 97\% completeness with
near-zero redundancy.
These results suggest that, in long-horizon procedurally constrained legal inquiry, reliable task completion is not an emergent property of probabilistic language models, but requires explicit structural enforcement to be guaranteed.

The remainder of this paper is organized as follows:
Section~2 reviews related work, Section~3 presents the theoretical framework,
Sections~4--5 describe the experimental setup and results, Section~6 discusses
implications and limitations, and Section~7 concludes.

\section{Related Work}

\subsection{Behavioral Consistency and Context Drift}
Prior work has extensively studied \textbf{context drift} in multi-turn LLM interactions, where models gradually deviate from initial goals, personas, or factual consistency over extended dialogue. To mitigate this phenomenon, soft techniques such as equilibria prompting, self-reflection, and internal consistency checks have been proposed to stabilize the model’s output distribution \cite{drift2025}. 
While effective in maintaining surface-level behavioral coherence, these approaches do not introduce verifiable notions of procedural progress and therefore cannot guarantee task completion in long-horizon, asymmetric settings.

\subsection{Task-Oriented Dialogue and State Tracking}
Task-oriented dialogue research has long emphasized explicit state tracking to ensure
goal completion, with datasets such as MultiWOZ serving as foundational benchmarks
\cite{budzianowski2018multiwoz}. Although recent LLM-based systems increasingly replace
symbolic dialogue states with implicit representations, surveys indicate that
\textbf{goal displacement} and long-horizon instability remain unresolved as task
complexity grows \cite{survey_llm_dialogue2024}.

Unlike prior task-oriented dialogue systems that track progress via turn-based
slot filling, our setting defines procedural state over the monotonic accumulation
of verified information units, rendering progress independent of conversational order.

\subsection{Structural Control and Neuro-Symbolic Integration} 
Recent surveys on computational argumentation identify long-horizon logical inconsistency as a central open challenge for neural dialogue systems, motivating the integration of explicit structural control \cite{castagna2024}. Complementary empirical studies further demonstrate that purely probabilistic language models suffer from exponential degradation in long-horizon execution, even under sophisticated prompting or agentic formulations \cite{samiei2026, illusion2025}. More recently, system-level analyses show that internal mechanisms such as multi-task training, fine-tuning, or agent memory fail to guarantee convergence without explicit external state enforcement, often resulting in premature termination or implicit completion heuristics \cite{samiei2026}. Our work addresses this gap by operationalizing monotonic and verifiable procedural
progress through a Soft-FSM architecture tailored to legal cross-examination.

\subsection{Legal Agents and Procedural Simulation}
Recent legal simulation frameworks such as AgentCourt demonstrate that LLM-based agents can sustain extended courtroom interactions through explicit role separation and workflow control \cite{agentcourt2024}. 
However, progress in these systems is defined over procedural stages rather than monotonic information accumulation, allowing termination without guaranteeing informational convergence. 
Our work targets this orthogonal failure mode by enforcing externally verifiable progress over accumulated Key Information Units in legal cross-examination.

\section{Theoretical Framework: Asymmetric Inquiry and Long-Horizon Non-Convergence}
\subsection{Structural Asymmetry Between Inquirer and Respondent}

We treat long-horizon inquiry failure not as a mere optimization issue, but as a \textbf{structural non-convergence problem}. In long-horizon legal interrogation, the inquirer and respondent play fundamentally asymmetric roles. The respondent performs \textit{reactive generation}; unlike the inquirer, their responses constitute informational inputs but lack the agency to dictate procedural state transitions. In contrast, the inquirer must make high-order strategic decisions—such as selecting the inquiry focus and determining stage transitions—under incomplete information. The static inquiry graph assumed in this formulation constitutes a special case of a more general setting where the inquiry structure may be dynamically updated in response to testimony.

This asymmetry induces a high-dimensional sequential decision problem in which errors made by the inquirer structurally constrain the set of subsequently reachable states. As a result, early procedural misjudgments propagate forward, leading to path-dependent failures that cannot be corrected through local linguistic coherence alone. Consequently, maintaining task commitment over long horizons
cannot be reliably ensured by probabilistic language generation alone.

\subsection{Problem Formulation: Inquiry as DAG Traversal}

Motivated by the structural asymmetry discussed above, we model procedural control as an external finite-state abstraction.We argue that some form of explicit procedural structure is required to enforce the monotonic and DAG-constrained nature of legal inquiry when procedural correctness is required to be guaranteed. While multiple implementations may satisfy this requirement, external finite-state control provides a concrete and verifiable mechanism, which we adopt in this work.

We formalize legal interrogation not as open-ended dialogue, but as a state-traversal problem over a \textbf{Directed Acyclic Graph (DAG)}. Let $S$ be the procedural state space defined by tuples $s_t = (K_t, \phi_t)$, where $K_t \subseteq \mathcal{K}_{target}$ denotes the set of accumulated Key Information Units (KIUs), and $\phi_t$ represents the linguistic context.

Unlike general conversation, a valid legal inquiry must satisfy the \textbf{Procedural Monotonicity Constraint}: a transition is considered progressive if and only if the information set strictly increases, i.e., $|K_{t+1}| > |K_t|$. Under this constraint, the inquiry process induces a trajectory $\tau = (s_0, \dots, s_T)$ that reaches the completion state $K_T = \mathcal{K}_{\text{target}}$ through strictly monotonic transitions in the information space, thereby traversing the DAG structure of the case narrative.

It is important to distinguish between conversational sequence and informational state trajectory. In natural discourse, an inquirer may revisit earlier topics or details for clarification. Such conversational backtracking does not constitute a procedural cycle, as long as the accumulated information state continues to advance. Therefore, even when the dialogue surface form exhibits non-linear topic shifts, the underlying inquiry remains a monotonic traversal over the DAG defined on the information state.

\begin{figure}[tb]
\centering
\begin{tikzpicture}[
    state/.style={draw, rectangle, rounded corners, minimum width=1.4cm, minimum height=0.6cm, align=center, font=\scriptsize},
    arrow/.style={->, thick},
    looparrow/.style={->, thick, bend left=25},
    >=Stealth
]

\node[state] (s0) {$s_0$\\$|K|=0$};
\node[state, right=0.4cm of s0] (s1) {$s_1$\\$|K|=1$};
\node[state, right=0.4cm of s1] (s2) {$\dots$};
\node[state, right=0.4cm of s2] (smax) {$s_{K_{\max}}$\\Comp.};

\draw[arrow] (s0) -- (s1);
\draw[arrow] (s1) -- (s2);
\draw[arrow] (s2) -- (smax);

\node[above=0.4cm of s1.east, font=\tiny\bfseries, xshift=0.3cm] {Progress Path (Monotonic)};

\node[state, below=1.5cm of s1, xshift=0.5cm] (sa) {$s_a$\\$|K|=k$};
\node[state, right=0.6cm of sa] (sb) {$s_b$\\$|K|=k$};

\draw[looparrow] (sa) to (sb);
\draw[looparrow] (sb) to (sa);

\node[below=0.6cm of $(sa.south)!0.5!(sb.south)$, font=\tiny\bfseries] {Stagnation Region (Cyclic)};

\draw[arrow, dashed] (s1.south) -- node[right, font=\tiny, align=left]{procedural\\misjudgment} (sa.north);

\end{tikzpicture}
\caption{
\textbf{Structural divergence between progressive and stagnating state transitions.}
State advancement is defined by information gain in $K$, independent of conversational order.
In the absence of external control, per-turn procedural misjudgments cause inquiry trajectories
to deviate from the required \textbf{Directed Acyclic Graph (DAG)} of information states and
collapse into a \emph{Stagnation Region}—a cyclic attractor where behavioral consistency is
preserved but informational progress stalls (i.e., $|K_{t+1}| = |K_t|$).
In contrast, Soft-FSM enforces traversal along a monotonic \emph{Progress Path} by permitting
state transitions only when the information set strictly increases
(i.e., $|K_{t+1}| > |K_t|$), ensuring monotonic advancement toward the terminal state $s_{K_{\max}}$.
}

\label{fig:stagnation}
\end{figure}
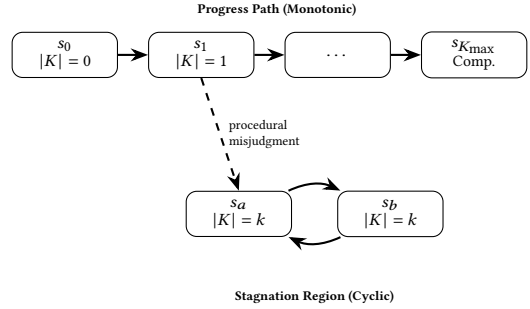
\subsection{The Complexity Cliff: Cyclic Stagnation}

We identify a fundamental mismatch between the procedural requirements of long-horizon inquiry—namely, monotonic traversal over a DAG of information states—and the inductive bias of large language models. Autoregressive LLMs optimize next-token likelihood $P(\phi_{t+1} \mid \phi_t)$, favoring local linguistic coherence over global procedural progress, a limitation that has been repeatedly observed in long-horizon execution settings~\cite{illusion2025,drift2025}.

This mismatch gives rise to the \textbf{Complexity Cliff}, where surface-level conversational fluency is preserved while underlying task progress collapses. Intuitively, a \textbf{stagnation region} can be understood as an equivalence class of inquiry states in which the linguistic context evolves but the accumulated information set remains unchanged:
\begin{equation}
\mathcal{R}_{\text{stagnant}} =
\{ (s_t, s_{t+1}) \mid \phi_{t+1} \neq \phi_t \land |K_{t+1}| = |K_t| \}.
\end{equation}

Prior work has shown that, in the absence of explicit external control, such non-progressive loops tend to persist and intensify over long interaction horizons~\cite{samiei2026}. As a result, conversational stability—often measured by low perplexity or stylistic consistency—can mask a complete lack of procedural advancement, leading to failures that are difficult to detect from language output alone~\cite{drift2025}.

\subsection{Soft-FSM Architecture: State-Grounded Decoding}

While classical dialogue FSMs enforce stage transitions based on turn-level slots,
Soft-FSM grounds procedural transitions exclusively in verifiable information gain.
To operationalize the finite-state abstraction introduced above, \textbf{Soft-FSM}
externalizes procedural state transitions into an explicit control layer.
Each inquiry stage is associated with mandatory Key Information Units (KIUs),
and progression is permitted only when the required information has been verified.

From a structural perspective, the underlying \textbf{finite-state topology}
enforces monotonic advancement by construction: transitions that do not increase
the information state are disallowed, effectively pruning non-progressive cycles
and restricting inquiry trajectories to DAG-consistent paths.
This form of externalized control has been shown to be effective for maintaining task commitment in long-horizon agentic systems. ~\cite{samiei2026,plan_act2025}.

Soft-FSM realizes this control through a \textbf{neuro-symbolic interface} that
decouples linguistic generation from procedural decision-making.
At each turn, the language model generates locally appropriate utterances
conditioned on the current state and unmet KIUs, while global state transitions
are governed by deterministic predicates rather than model confidence or
self-evaluation. This \textbf{state-grounded decoding} constrains the procedural
effects of language outputs without sacrificing linguistic flexibility,
addressing known limitations of prompt-based and purely probabilistic control
strategies~\cite{toolformer2023,survey_llm_dialogue2024}.

\section{Experimental Setup}
This section describes the experimental setup used to evaluate Soft-FSM. Our goal is to systematically compare whether different interrogation architectures can reliably maintain procedural progress and task completion in long-horizon, structured legal inquiry tasks. This choice is intended to isolate the structural behavior of the inquiry controller from confounds introduced by automated extraction.

\subsection{Cases and Task Definition}
We evaluate our approach on three real-world criminal homicide cases drawn from the lay judge system, representing increasing levels of complexity:
\begin{itemize}
    \item \textbf{Case A (Simple/Confession):} A single defendant fully confesses; the case is linear with minimal factual dispute.
    \item \textbf{Case B (Complex/Disputed Intent):} The defendant admits the act but denies intent, requiring inquiry into motive and sentencing factors.
    \item \textbf{Case C (Abstract/Denial):} The defendant denies the offense or involves multiple accomplices and complex forensic evidence.
\end{itemize}
For each case, we manually construct a structured target schema consisting of over 40 \textbf{Key Information Units (KIUs)}. The task is considered complete only when all KIUs are successfully elicited.

\subsection{Compared Methods}
We compare four interrogation strategies using \texttt{Gemma-3-27B-it} as the base model:
\begin{itemize}
    \item \textbf{V1 (Pure LLM):} Instructed only with the prosecutor role and a high-level goal.
    \item \textbf{V2 (Stage-Prompted):} Provided with a full procedural SOP in the system prompt, but no external state tracking is enforced.
    \item \textbf{V3 (Soft-FSM - Ours):} An external finite-state controller governs stages and KIU progress; the LLM is used solely for question generation.
    \item \textbf{V4 (Equilibria-Prompted):} Inspired by \cite{drift2025}, the model performs continuous self-checking against goals without external deterministic control.
    We omit V4 from aggregate plots for clarity, as its performance closely mirrors V1 with slightly improved stability but no improvement in completeness.
\end{itemize}

\subsection{Oracle Witness and Metrics}
To isolate procedural behavior from response variability, we employ a deterministic oracle witness grounded in the official written court judgments of each case. At each turn, the oracle provides a single factual response drawn exclusively from the corresponding judgment text, ensuring that all ground-truth information is fixed and externally verifiable.

Examination proceeds in a strict question--answer format: the inquirer asks one question per turn, and the oracle returns the minimal judgment-supported answer without inference or elaboration. A Key Information Unit (KIU) is considered successfully elicited if and only if the oracle response contains the corresponding factual element present in the judgment.

We report four metrics: (1) \textbf{Completeness}, defined as the proportion of KIUs successfully elicited from the judgment; (2) \textbf{Redundancy}, the fraction of questions targeting already-filled KIUs; (3) \textbf{Unknown Rate}, the fraction of questions that do not correspond to any KIU in the judgment; and (4) \textbf{Stability}, measured as the standard deviation of completeness across repeated runs.
\FloatBarrier
\section{Experimental Results}

\subsection{Overall Performance: The Complexity Cliff}
\begin{figure}[!ht]
\centering
\begin{tikzpicture}
\begin{axis}[
    width=\linewidth,
    height=6cm,
    xlabel={Case Complexity},
    ylabel={Completeness (\%)},
    xmin=0.5, xmax=3.5,
    ymin=0, ymax=110,
    xtick={1,2,3},
    xticklabels={Case A\\(Simple), Case B\\(Complex), Case C\\(Abstract)},
    xticklabel style={align=center, font=\small},
    legend pos=south west,
    legend style={font=\tiny, row sep=-0.5pt},
    ymajorgrids=true,
    grid style=dashed,
    title={\textbf{Completeness vs. Case Complexity}},
    title style={font=\small, yshift=-5pt}
]

\addplot[
    color=blue,
    mark=square*,
    thick,
    error bars/.cd,
    y dir=both, y explicit
] coordinates {
    (1, 97.2) +- (0, 3.8)
    (2, 97.2) +- (0, 3.8)
    (3, 98.6) +- (0, 3.1)
};
\addlegendentry{V3: Soft-FSM (Ours)}

\addplot[
    color=red,
    mark=*,
    thick,
    dashed,
    error bars/.cd,
    y dir=both, y explicit
] coordinates {
    (1, 67.4) +- (0, 13.6)
    (2, 38.6) +- (0, 22.2)
    (3, 35.8) +- (0, 12.9)
};
\addlegendentry{V1: Pure LLM}

\addplot[
    color=gray,
    mark=triangle*,
    thick,
    dotted,
    error bars/.cd,
    y dir=both, y explicit
] coordinates {
    (1, 33.5) +- (0, 13.2)
    (2, 19.1) +- (0, 3.8)
    (3, 20.5) +- (0, 6.9)
};
\addlegendentry{V2: Stage Prompt}

\end{axis}
\end{tikzpicture}
\caption{\textbf{The Complexity Cliff.}
Soft-FSM (Ours) maintains near-perfect completeness ($>97\%$) regardless of task complexity,
while baseline methods degrade sharply as procedural depth increases.
V3 maintains near-zero variance across all cases, demonstrating structural robustness.
Error bars denote standard deviation ($N=5$).}
\label{fig:complexity_cliff}
\end{figure}
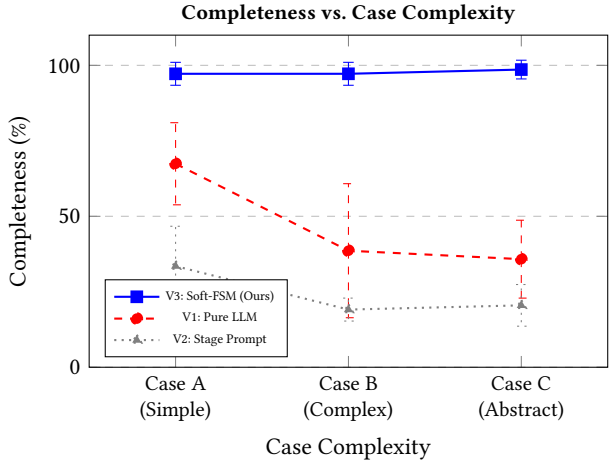

Figure~\ref{fig:complexity_cliff} illustrates the average completeness across cases. Soft-FSM consistently achieves $97$--$99\%$ completeness with near-zero variance, independent of case difficulty. 

\begin{table}[!ht]
\centering
\caption{Performance comparison (mean $\pm$ std, $N=5$). V3 (Ours) demonstrates structural robustness in long-horizon inquiry tasks.}
\label{tab:main_results}
\resizebox{\columnwidth}{!}{
\begin{tabular}{llccc}
\toprule
\textbf{Case} & \textbf{Method} & \textbf{Comp. (\%)} & \textbf{Redun. (\%)} & \textbf{Unk. (\%)} \\
\midrule
\textbf{Case A} & V1: Pure LLM & $67.4 \pm 13.6$ & $6.4 \pm 4.0$ & $22.5 \pm 17.2$ \\
(Simple)        & V2: Stage Prompt & $33.5 \pm 13.2$ & $38.8 \pm 14.1$ & $30.5 \pm 27.8$ \\
                & \textbf{V3: Soft-FSM (Ours)} & $\mathbf{97.2 \pm 3.8}$ & $\mathbf{0.0 \pm 0.0}$ & $\mathbf{8.0 \pm 11.0}$ \\
\midrule
\textbf{Case B} & V1: Pure LLM & $38.6 \pm 22.2$ & $7.1 \pm 10.5$ & $50.5 \pm 29.3$ \\
(Complex)       & V2: Stage Prompt & $19.1 \pm 3.8$ & $76.3 \pm 11.2$ & $14.5 \pm 2.7$ \\
                & \textbf{V3: Soft-FSM (Ours)} & $\mathbf{97.2 \pm 3.8}$ & $\mathbf{0.0 \pm 0.0}$ & $\mathbf{8.0 \pm 11.0}$ \\
\midrule
\textbf{Case C} & V1: Pure LLM & $35.8 \pm 12.9$ & $0.9 \pm 2.0$ & $60.5 \pm 15.5$ \\
(Abstract)      & V2: Stage Prompt & $20.5 \pm 6.9$ & $66.2 \pm 20.1$ & $24.0 \pm 9.5$ \\
                & \textbf{V3: Soft-FSM (Ours)} & $\mathbf{98.6 \pm 3.1}$ & $\mathbf{0.0 \pm 0.0}$ & $\mathbf{4.0 \pm 8.9}$ \\
\bottomrule
\end{tabular}
}
\end{table}
Consistent with the failure probability bound $P(\text{failure}) \ge 1 - (1-\epsilon)^n$, V1 exhibits a clear Complexity Cliff: completeness drops from 67.4\% in Case A to 35.8\% in Case C as task depth $n$ increases. This pattern indicates that, in the absence of structural enforcement, even small per-turn procedural misjudgments accumulate over long horizons and manifest as global failures in task completion. We additionally observe that V4 (Equilibria-Prompted) exhibits performance comparable to the pure LLM baseline, failing to mitigate the sharp drop in completeness under increasing task complexity. This suggests that internal self-consistency mechanisms alone are insufficient to prevent the stagnation phenomena underlying the Complexity Cliff.
\subsection{Behavioral Analysis}
Analysis reveals that V2 (Stage-Prompted) frequently enters paraphrasing loops. While the model maintains \textbf{behavioral consistency}, it fails to achieve \textbf{procedural progress}, resulting in high redundancy ($>66\%$). In contrast, Soft-FSM ensures monotonic information accumulation.
\subsection{Summary}
These results highlight a clear separation between behavioral coherence and procedural advancement in LLM-based legal inquiry. Baseline methods maintain linguistic stability but consistently fail to achieve informational convergence, exhibiting the Complexity Cliff in complex cases (completeness $<40\%$). In contrast, Soft-FSM enforces monotonic KIU accumulation, yielding near-complete coverage ($>97\%$) with minimal redundancy across all cases.
This empirical pattern supports our claim that long-horizon task completion in procedurally constrained domains is reliably achieved when progress is explicitly enforced through deterministic state transitions, rather than relying on probabilistic alignment alone.

\section{Discussion}
\subsection{Task Commitment as a Structural Gap}

Our results confirm the \textbf{Complexity Cliff}: baseline methods preserve
linguistic coherence but structurally collapse into non-advancing inquiry loops,
reflecting the inductive bias of probabilistic models toward local plausibility
over long-horizon progression~\cite{samiei2026,illusion2025}.

We distinguish \textit{procedural stagnation} from \textit{divergence}.
Stagnation corresponds to a verifiable halt in KIU accumulation, whereas divergence
manifests as unverifiable hallucination.
Notably, the incidence of stagnation is \emph{sensitive to KIU granularity}:
sparser information schemas increase the likelihood of encountering non-progressive
regions even under stable linguistic behavior.

Under Soft-FSM, such failures remain structurally bounded.
The finite-state topology enforces monotonic information gain, ensuring that failures
manifest as detectable non-advancement rather than unconstrained generation.
This aligns with recent findings that externalized control provides a reliable mechanism for maintaining task commitment in long-horizon agentic systems~\cite{samiei2026,plan_act2025}.

\subsection{Why Soft Corrections Fail to Guarantee Monotonic Progress}

Recent approaches to mitigating drift, such as equilibria prompting~\cite{drift2025}, operate primarily at the linguistic level. While they guide model reasoning or self-correction, they do not impose hard constraints on procedural state advancement. As the probability of procedural misjudgment at each interaction step remains non-zero, such errors accumulate over long horizons, trapping systems in \textbf{stagnation regions}—cyclic trajectories in which behavioral consistency is preserved without progress in the underlying information state.

This limitation aligns with prior agent studies showing that reliable long-horizon behavior requires \emph{externalized} planning and execution control~\cite{plan_act2025,toolformer2023}. In our setting, monotonic progress can be enforced by an explicit finite-state control layer, which Soft-FSM operationalizes via deterministic stage transitions, preventing probabilistic deviations from escalating into global procedural failure. \textbf{Importantly, we argue that such external control 
provides a highly reliable mechanism observed to prevent such probabilistic deviations from escalating into global failure.}

\subsection{Intelligence as Goal-Directed Navigation}
A key insight is that linguistic fluency does not guarantee procedural progress: models can sustain behavioral stability while trapped in stagnation loops. In constrained legal inquiry, intelligence emerges as sustained monotonic navigation through a DAG-defined information state space~\cite{sutton1999options}.

Soft-FSM realizes this by grounding transitions solely in verifiable KIU gain, decoupled from conversational order. This reframes reliable completion as goal-directed traversal under structural constraints~\cite{alur2015correct}, rather than an emergent outcome of next-token prediction.

\subsection{Limitations and Scope}
We note two limitations. First, expert-defined KIU schemas are used to isolate the structural behavior of the inquiry controller from noise introduced by automated extraction; integrating learned information extraction is left for future work. Second, our use of an Oracle witness focuses on failures arising from the inquirer’s internal logic. In real-world adversarial cross-examination, the inquiry structure may evolve dynamically, yet monotonic information gain remains a necessary condition for guaranteeing the absence of procedural stagnation, regardless of whether the inquiry graph is static or dynamically computed.

\section{Conclusion}

This work addresses a fundamental \textbf{topological mismatch} between the cyclic inductive
biases of large language models and the monotonic requirements of long-horizon legal inquiry.
We identify the \textbf{Complexity Cliff}—a structural failure mode in which models preserve
behavioral coherence while falling into procedural stagnation—and introduce \textbf{Soft-FSM}
as a principled mechanism for mitigating this mismatch through explicit external state control.
Empirical results show that while purely probabilistic baselines collapse in complex inquiry
settings (completeness $<40\%$), the proposed neuro-symbolic approach consistently achieves
near-complete ($>97\%$) information coverage by pruning non-progressive transitions.

Crucially, our findings demonstrate that behavioral stability is orthogonal to procedural
progress. In DAG-structured inquiry tasks, the absence of observable drift does not imply
advancement toward completion. From a structural perspective, \textbf{
explicit external state enforcement provides a principled and verifiable mechanism for 
guaranteeing deterministic goal completion under procedural constraints.} This reframes external state control as a
means of enforcing procedural commitment rather than optimizing linguistic behavior.

More broadly, our results suggest that for complex legal inquiry, purely probabilistic alignment
is insufficient to guarantee reliable task execution. Instead, externalized topological
enforcement offers a principled mechanism for bridging the gap between linguistic fluency and
deterministic goal completion. We do not claim this to be the only possible realization of such structure, but demonstrate it as a concrete and effective instantiation.

\bibliographystyle{ACM-Reference-Format}
\bibliography{references}

@article{castagna2024,
  title   = {Computational Argumentation-based Chatbots: A Survey},
  author  = {Castagna, Federico and K{\"o}kciyan, Nadin and Sassoon, Isabel and Parsons, Simon and Sklar, Elizabeth},
  journal = {Journal of Artificial Intelligence Research},
  volume  = {80},
  pages   = {1271--1310},
  year    = {2024},
  doi     = {10.1613/jair.1.15407},
  url     = {https://jair.org/index.php/jair/article/view/15407}
}

@article{illusion2025,
  title   = {The Illusion of Diminishing Returns: Measuring Long-Horizon Execution in {LLMs}},
  author  = {Sinha, Akshit and Arun, Arvindh and Goel, Shashwat and Staab, Steffen and Geiping, Jonas},
  journal = {arXiv preprint arXiv:2509.09677},
  year    = {2025},
  url     = {https://arxiv.org/abs/2509.09677}
}

@article{samiei2026,
  title   = {The Illusion of Procedural Reasoning: On the Failure of Internal Control in Long-Horizon Agentic Language Models},
  author  = {Samiei, Amir and others},
  journal = {arXiv preprint},
  year    = {2026}
}

@article{drift2025,
  title   = {Drift No More? Context Equilibria in Multi-Turn {LLM} Interactions},
  author  = {Cirrincione, Luca and Hanina, Talal and Vignali, Ugo and D'Auria, Daniela and Dubrawski, Artur},
  journal = {arXiv preprint arXiv:2510.07777},
  year    = {2025},
  url     = {https://arxiv.org/abs/2510.07777}
}

@inproceedings{budzianowski2018multiwoz,
  title     = {{MultiWOZ}: A Large-Scale Multi-Domain Wizard-of-{Oz} Dataset for Task-Oriented Dialogue Modelling},
  author    = {Budzianowski, Pawe{\l} and Wen, Tsung-Hsien and Tseng, Bo-Hsiang and Casanueva, I{\~n}igo and Ultes, Stefan and Ramadan, Osman and Ga{\v{s}}i{\'c}, Milica},
  booktitle = {Proceedings of the 2018 Conference on Empirical Methods in Natural Language Processing (EMNLP)},
  year      = {2018},
  url       = {https://aclanthology.org/D18-1547}
}

@article{survey_llm_dialogue2024,
  title   = {A Survey on Recent Advances in {LLM}-Based Multi-turn Dialogue Systems},
  author  = {Yi, Zihao and Ouyang, Jiarui and Xu, Zhe and Liu, Yuwen and Liao, Tianhao and Luo, Haohao and Shen, Ying},
  journal = {arXiv preprint arXiv:2402.18013},
  year    = {2024},
  url     = {https://arxiv.org/abs/2402.18013}
}

@article{plan_act2025,
  title   = {Plan-and-Act: Improving Planning of Agents for Long-Horizon Tasks},
  author  = {Erdogan, Lutfi Eren and Lee, Nicholas and Kim, Sehoon and Moon, Suhong and Furuta, Hiroki and Anumanchipalli, Gopala and Keutzer, Kurt and Gholami, Amir},
  journal = {arXiv preprint arXiv:2503.09572},
  year    = {2025},
  url     = {https://arxiv.org/abs/2503.09572}
}

@article{toolformer2023,
  title   = {{Toolformer}: Language Models Can Teach Themselves to Use Tools},
  author  = {Schick, Timo and Dwivedi-Yu, Jane and Dess{\`i}, Roberto and Raileanu, Roberta and Lomeli, Maria and Zettlemoyer, Luke and Cancedda, Nicola},
  journal = {arXiv preprint arXiv:2302.04761},
  year    = {2023},
  url     = {https://arxiv.org/abs/2302.04761}
}

@article{agentcourt2024,
  title   = {{AgentCourt}: Simulating Court with Adversarial Evolvable Lawyer Agents},
  author  = {Chen, Guhong and Fan, Liyang and Gong, Zihan and Xie, Nan and Li, Zixuan and Liu, Ziqiang and Li, Chengming and Qu, Qiang and Alinejad-Rokny, Hamid and Ni, Shiwen and Yang, Min},
  journal = {arXiv preprint arXiv:2408.08089},
  year    = {2024},
  url     = {https://arxiv.org/abs/2408.08089}
}

@article{alur2015correct,
  title   = {Correct-by-Construction Synthesis of Reactive Systems},
  author  = {Alur, Rajeev and others},
  journal = {Communications of the ACM},
  volume  = {58},
  number  = {4},
  pages   = {68--78},
  year    = {2015},
  doi     = {10.1145/2699415}
}

@article{sutton1999options,
  title   = {Between {MDPs} and Semi-{MDPs}: A Framework for Temporal Abstraction},
  author  = {Sutton, Richard S and Precup, Doina and Singh, Satinder},
  journal = {Artificial Intelligence},
  volume  = {112},
  number  = {1--2},
  pages   = {181--211},
  year    = {1999}
}

\end{document}